\documentclass{article}
\usepackage{multirow}
\usepackage{booktabs}
\usepackage[english]{babel}
\usepackage[letterpaper,top=2cm,bottom=2cm,left=3cm,right=3cm,marginparwidth=1.75cm]{geometry}
\usepackage{amsmath}
\usepackage{graphicx}
\usepackage[colorlinks=true, allcolors=blue]{hyperref}

\title{FIPGNet:Pyramid grafting network with feature interaction strategies }
\author{Ziyi Ding, Like Xin*}

\begin{document}
\maketitle
\begin{abstract}
Salient object detection is designed to identify the objects in an image that attract the most visual attention.Currently, the most advanced method of significance object detection adopts pyramid grafting network architecture.However, pyramid-graft network architecture still has the problem of failing to accurately locate significant targets.We observe that this is mainly due to the fact that current salient object detection methods simply aggregate different scale features, ignoring the correlation between different scale features.To overcome these problems, we propose a new salience object detection framework(FIPGNet),which is a pyramid graft network with feature interaction strategies.Specifically, we propose an attention-mechanism based feature interaction strategy (FIA) that innovatively introduces spatial agent Cross Attention (SACA) to achieve multi-level feature interaction, highlighting important spatial regions from a spatial perspective, thereby enhancing salient regions.And the channel proxy Cross Attention Module (CCM), which is used to effectively connect the features extracted by the backbone network and the features processed using the spatial proxy cross attention module, eliminating inconsistencies.Finally, under the action of these two modules, the prominent target location problem in the current pyramid grafting network model is solved.Experimental results on six challenging datasets show that the proposed method outperforms the current 12 salient object detection methods on four indicators.
\end{abstract}

\section{Introduction}
Salient object detection (SOD) is designed to identify the most visually obvious areas.It has developed rapidly with the help of deep learning methods and, as a pre-processing step, is widely used in a variety of computer vision tasks, Examples include object detection \cite{ren2013region}, semantic segmentation\cite{sun2020mining} , image understanding\cite{zhu2014unsupervised} , object discovery \cite{karpathy2013object}, image classification \cite{wu2024tai++}, 4D significance detection\cite{wang2019deep}  quality assessment of synthetic images without reference\cite{wang2019no} and open environment\cite{yang2024robust}.In addition, SOD can also be extended to multi-modal tasks, including RGB-D SOD\cite{li2023mutual}, RGB-T SOD\cite{chen2023light}, and optical field SOD\cite{zhang2017amulet}.

Although great progress has been made, most of the advanced salient object detection methods adopt the structure of encoding and decoding. As a result, salient objects cannot be accurately detected, as shown in Figure 1. Yuan et al \cite{yuan2023m} shows that low-level features contain more non-significant information and background noise, while high-level features contain more significant information. How to effectively interact encoder features and improve the correlation among features to highlight the significant information still needs attention.

To solve the above problems, some methods integrate the features of different layers layer by layer, as shown in Figure 2 (a).FCN\cite{long2015fully} connects the corresponding level of features in the encoder to the decoder through the transport layer, and directly splice all the features from the deep and shallow layers to integrate the multi-level features.GateNet\cite{zhao2020suppress} uses a gate mechanism to balance the contribution of each encoder block and reduce non-significant information. However, simply splicing all the layers without considering the importance of interaction is not an effective way to merge.Other methods take into account the interaction between features, but the feature interaction exists only among a few layers and has limitations, as shown in Figure 2 (b).BMPM\cite{zhang2018bi} uses a bidirectional structure to transfer information between features at different levels, with high-level semantic information passed to the shallow layer, and low-level spatial details contained in the shallow layer features passed to the opposite direction, so that semantic information and detail information are inserted into each level.However, single-level features can only describe information at a specific scale, and in the process of top-down, the ability of shallow features to represent details is weakened.To take advantage of multiple layers of features, some methods combine multiple layers of features in a fully connected manner or heuristically, as shown in Figure 2(c).Amulet integrates multi-level feature maps into multiple resolutions, while combining coarse semantics and fine details to adaptively learn to combine these feature maps at each resolution. However, the integration of too many features, lack of balance between different scale features, easy to lead to high computing costs, noise and thus affect the decoder information recovery.In addition, some methods are proposed to make better use of multi-level features by means of mutual learning, so as to avoid the interference caused by the differences of features at different scales, as shown in Figure 2(d).MINet\cite{pang2020multi} effectively integrates contextual information from adjacent layer features and learns multi-scale features from a single convolutional block by training two different interactive branches. Through interactive learning, branches can more flexibly integrate information from other branches.However, this method can only integrate the feature information of adjacent layers, and redundant information is inevitably generated through interactive learning.

Inspired by the work of Wang et al \cite{wang2023narrowing} and Han et al \cite{pang2020multi}, we propose an attention-mechanism-based feature interaction strategy to better learn the spatial correlation between multi-scale features, as shown in Figure 2(e).Specifically, the method first connects all the layer features of the encoder side in series along the spatial dimension, and then sends them together into the spatial agent Cross attention block (SACA) for attention operation, so as to determine the important regions on all the scale features.Since the significant region enhancement features obtained at this stage cannot be directly upsampled and decoded, we input them together with the original features obtained by the encoder into the channel agent Cross Attention module (CCM), using which the significant region enhancement features can be calibrated to the original features.Among them, spatial agent cross attention and channel agent cross attention realize more efficient attention operation.

The main contributions of this paper are as follows:
\begin{itemize}
    \item We propose a feature interaction strategy based on attention mechanism(FIA).Innovatively introduces spatial agent Cross Attention (SACA) to achieve multi-level feature interaction, highlighting important spatial regions from a spatial perspective, thereby enhancing salient regions.And the channel proxy Cross Attention Module (CCM), which is used to effectively connect the features extracted by the backbone network and the features processed using the spatial proxy cross attention module, eliminating inconsistencies.
\end{itemize}
\begin{itemize}
    \item We construct a new pyramidal graft network with feature interaction strategies(FIPGNet).We compared the proposed method with 12 state-of-the-art SOD methods on 6 datasets. The experimental results show that our method is superior to other methods.
\end{itemize}

\begin{figure}
\centering
\includegraphics[width=0.75\linewidth]{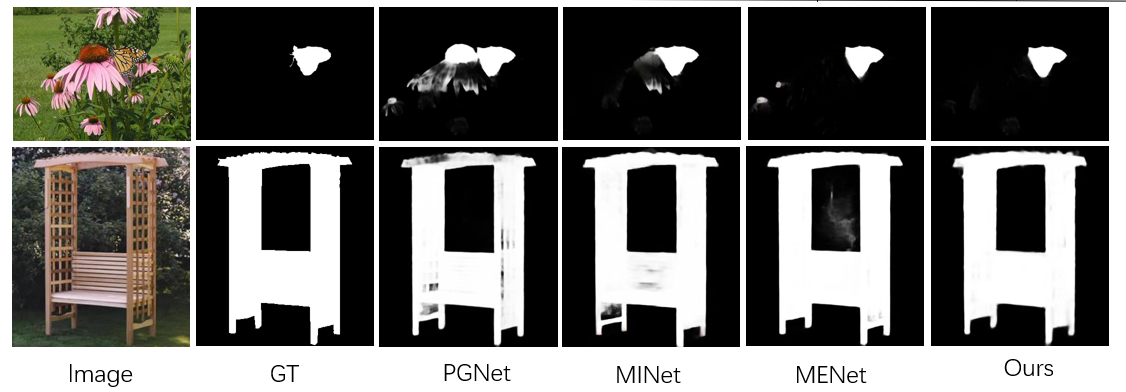}
\caption{The comparison between our method and other methods in salient object detection }
\label{fig:1}
\end{figure}

\section{Related Work}

\begin{figure}
\centering
\includegraphics[width=0.75\linewidth]{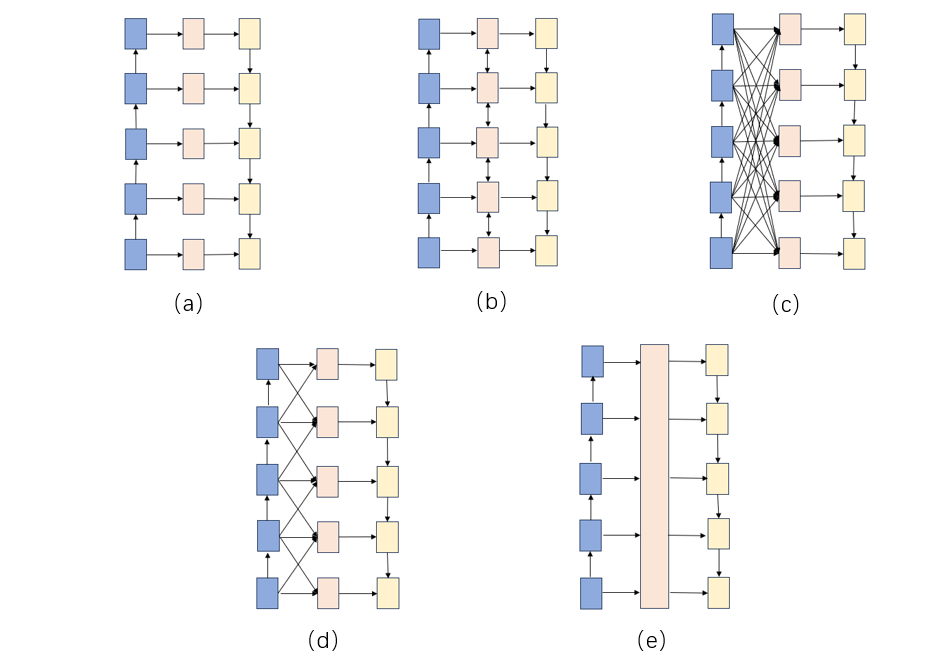}
\caption{Illustration of different architectures. Blue blocks, pink blocks and yellow blocks respectively denote the different convolutional blocks in the encoder, the transport layer and the decoder.}
\label{fig:2}
\end{figure}

\subsection{Salient Object Detection based on deep learning}

CNN-based methods are widely used in SOD and gain conmendable performance, which usually takes the pre-trained networks \cite{he2016deep} as the encoder, and most of the efforts are then to design an effective decoder to conduct multilevel feature aggregation. Most methods\cite{zhang2017amulet}, \cite{zhao2020suppress}, \cite{pang2020multi} use the U-shape  \cite{ronneberger2015u}based structures as the encoder-decoder architecture and progressively aggregate the hierarchical features for final prediction.

Transformer encoder-decoder architecture was first proposed by Vaswani et al \cite{vaswani2017attention}for natural language processing.Since the success of ViT \cite{vaswani2017attention} in image classification, more and more works have introduced Transformer architecture to computer vision tasks.SETR\cite{vaswani2017attention} and PVT \cite{wang2021pyramid} use ViT as the encoder for semantic segmentation. In SOD,VST \cite{liu2021visual} employs T2T-ViT \cite{yuan2021tokens} as the backbone and proposes an effective multi-task decoder for features in a sequence form.Yang et al\cite{yang2023towards} designed a context-aware converter with local global self-attention, effectively considering both long-term and short-term contexts. 

Previous Transformer-based methods are superior in locating and capturing the salient areas in the images, while the details at the local level could be ignored.Unlike previous CNN-based methods and Transformer-based methods, our FIPGNet  handle features extracted by Transformer and CNN graft pyramid encoder\cite{xie2022pyramid}and performs sequential processing of feature maps in the decoder.

\subsection{Multi-layer feature enhancement and interaction}

Multi-layer feature enhancement and interaction are crucial for gaining a high-resolution salient map, and numerous methods have thoroughly investigated them.GateNet\cite{ronneberger2015u} adopted the gate mechanism to balance the contribution of each encoder block and reduce non-salient information.MINet\cite{pang2020multi}  proposed an interactive learning method for multilevel features, aiming to minimize the discrepancies and improve the spatial coherence of multilevel features. ICON\cite{zhuge2022salient}  incorporated convolution kernels with different shapes to enhance the diversity of multilevel features. In addition, there has been some work to integrate attention mechanisms into the interaction of multi-scale features.Zhang et al\cite{liu2020picanet}propose a novel attention guided network which selectively integrates multi-level contextual information in a progressive manner.Tang et al\cite{tang2022hrtransnet}injected supplementary modes into the main modes by using global optimization and attention mechanisms to select and purify modes at the input level.Meng et al\cite{meng2024multiscale} introduced the modeling capability of the global group attention mechanism to make each header focus on different context information.The grid cross-attention mechanism is used to establish the relationship between visual features and text features at various scales. In addition,Other work uses attention mechanisms to deal with multi-modal features.Farid et al \cite{razzak2019integrated}used a dual-attention recurrent neural network to predict stress test measures based on dimensions and temporal characteristics selected from exogenous economic conditions and bank performance profiles.

To fully utilize multi-scale features,we propose a Feature Interaction Strategy based on attention mechanism(FIA).And Spatial Agent Cross Attention block (SACA) to achieve multi-level feature interaction, highlighting important spatial regions from a spatial perspective, thereby enhancing salient regions.And the channel proxy Cross Attention Module (CCM), which is used to effectively connect the features extracted by the backbone network and the features processed using the spatial proxy cross attention module, eliminating inconsistencies.

\section{Proposed method}

\subsection{Network Architecture}
Figure 3 shows the overall architecture of our FIPGNet network.Our network is built on the PGNet \cite{xie2022pyramid} architecture.The framework consists mainly of backbone network , FIA part (Section 3.2), CCM part  (Section 3.3), and  decoder.backbone and decoder follow the design of Xie et al\cite{xie2022pyramid}. 

\begin{figure}
\centering
\includegraphics[width=1\linewidth]{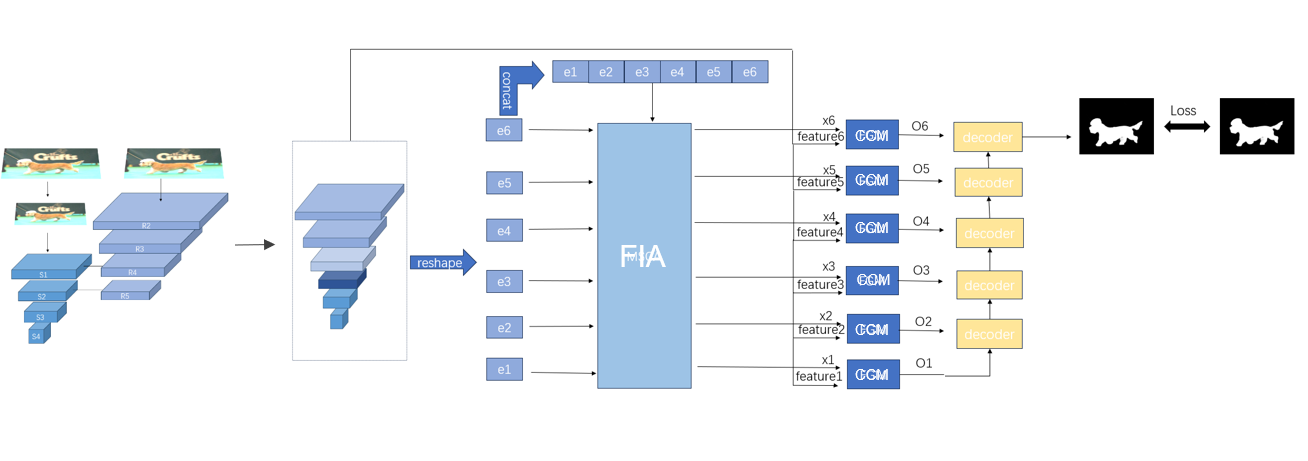}
\caption{Illustration of the proposed FIPGNet.The framework consists mainly of backbone network , FIA part , CCM, and decoder.}
\label{fig:3}
\end{figure}

\begin{figure}
\centering
\includegraphics[width=0.75\linewidth]{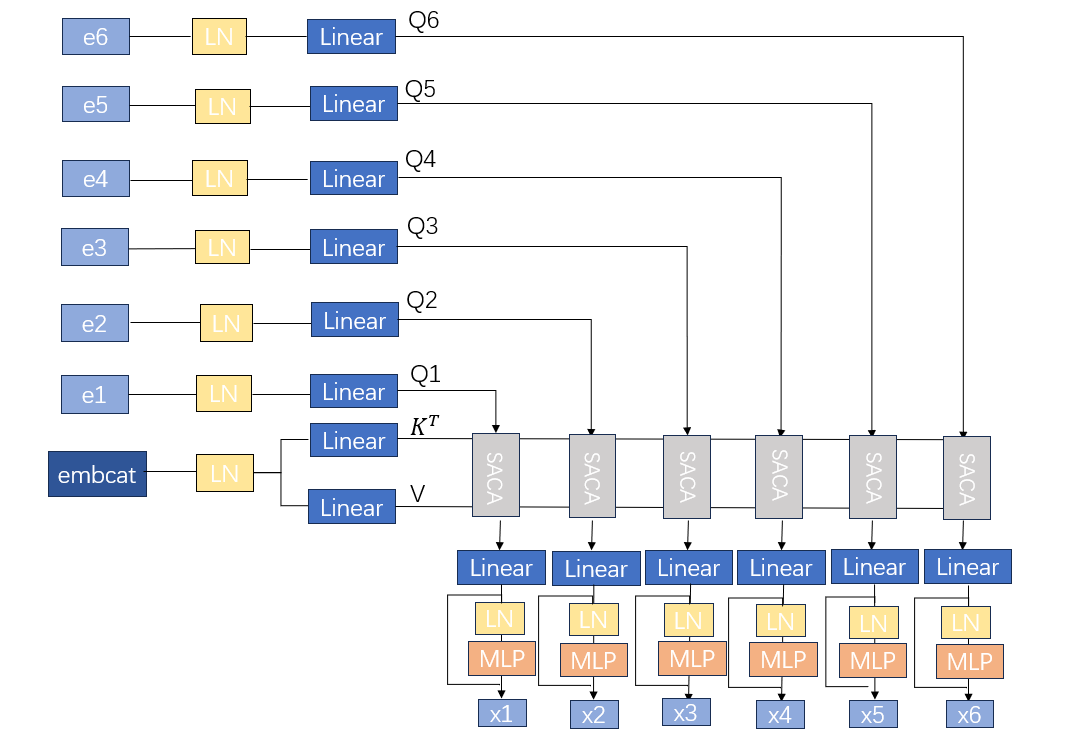}
\caption{Illustration of the learning procedure of the proposed FIA modules}
\label{fig:4}
\end{figure}

\subsection{FIA: Feature interaction strategies based on attention mechanisms}
Low-level features contain more non-salient information and background noise, while high-level features contain more salient information.On this basis, we design a multi-scale feature interaction strategy to enhance the spatial correlation between features and highlight the salient regions through the interaction between low-level features and high-level features.The FIA module consists of three phases: generating tokens, Spatial Agent Cross Attention (SACA), and MLP residual structure as shown in Figure 4.In the generation of tokens stage, the characteristics of different layers obtained by the backbone network are processed to generate tokens with the same space size and number of channels, which is convenient to process the characteristics in the spatial dimension in the subsequent stage.Spatial agent Cross Attention Blocks (SACA) perform attention operations on the processed features and learn the spatial correlation between features at different scales.The MLP residual structure refines the features obtained in the SACA stage.
\subsubsection{Generating Tokens}
We resize the image to a fixed size X and get six different scale features after input into the backbone network for processing.We followed the settings in Visual transformer to set the tokens size to 14,and the channel dimensions are all changed to C=128.Finally,We will concatenate the tokens ei from the features along the spatial dimension to get embcat.

\subsubsection{Spatial Agent Cross Attention (SACA)}
We input six ei features into SACA blocks respectively to conduct attention operations with embcat. The specific process is shown in Figure 5.First, we carry out ei pooling operation to get a smaller feature Query Agent Tokens AQ.AQ is regarded as the agent of feature ei and embcat is calculated with spatial cross-attention to get the agent feature.Then, we concat six Query Agent Tokens along the spatial dimension to get Key Agent Tokens.Key Agent Tokens act as the agents of feature embcat and then calculate the space cross attention with feature ei and Agent Features to get the output result.Following the work of Han et al, we defined Agent Bias to promote different agent vectors to focus on different locations in the picture, so as to make better use of location information.In addition, we also use a lightweight DWC as a diversity recovery module.

\begin{figure*}
\centering
\includegraphics[width=1\linewidth]{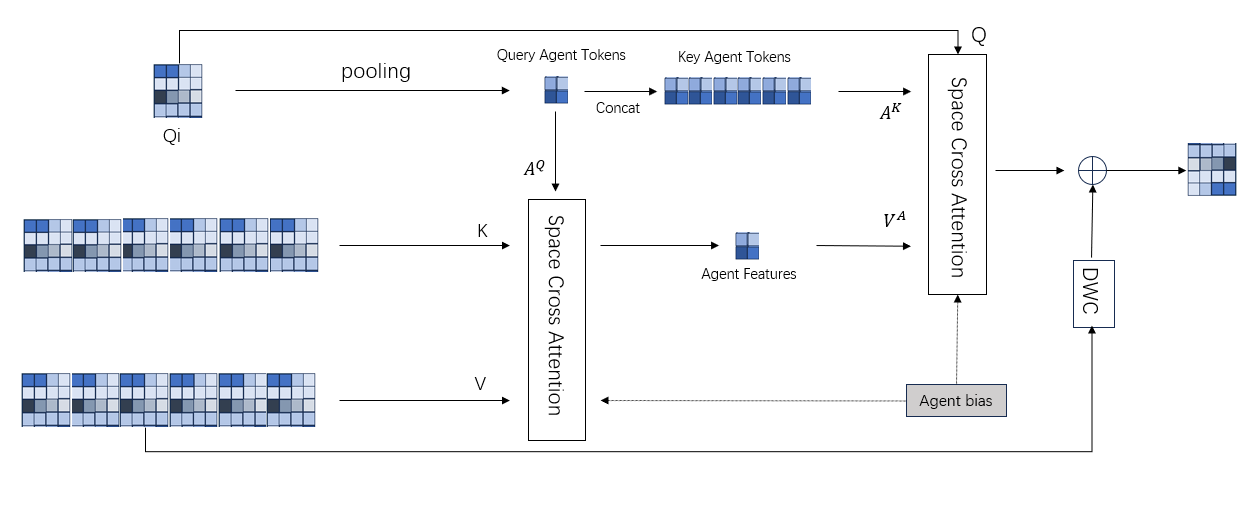}
\caption{Illustration of the learning procedure of the proposed SACA modules}
\label{fig:5}
\end{figure*}
\begin{figure*}
\centering
\includegraphics[width=1\linewidth]{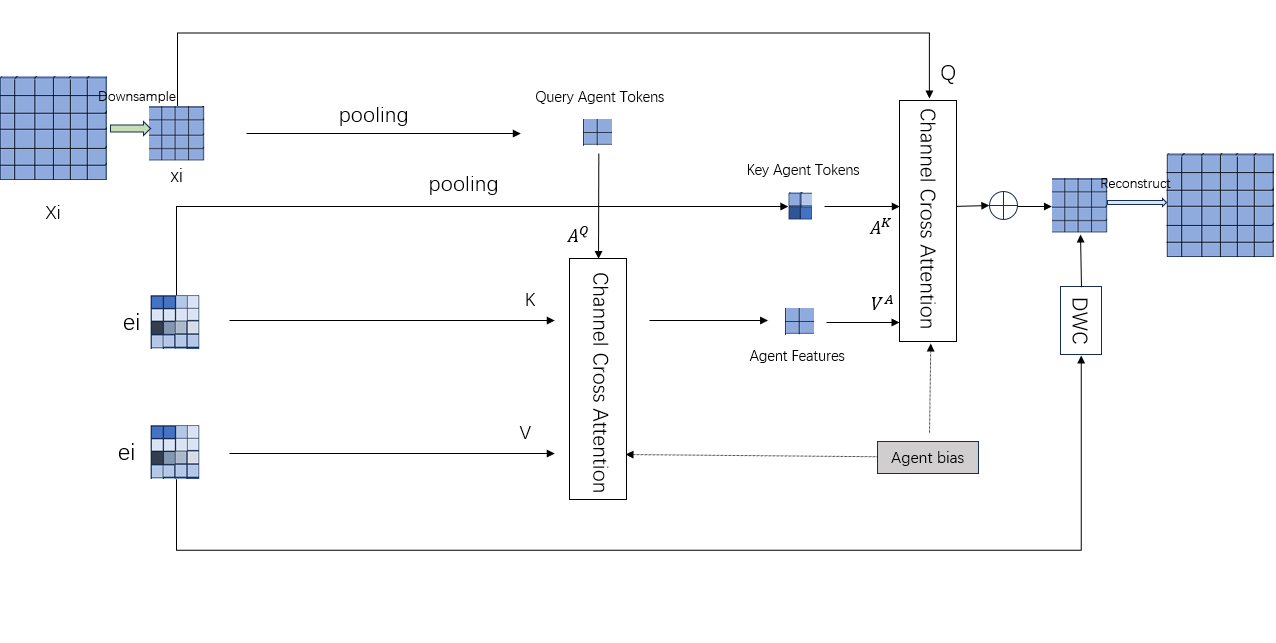}
\caption{Illustration of the learning procedure of the proposed CCA modules}
\label{fig:6}
\end{figure*}

\subsubsection{MLP Residual Structure}

Following SACA, MLPs with residual structure are developed to encode channel dependencies for refining the features from SACA. The output of each MLP is obtained as follow:

\begin{equation}  
    S_{i}= {S}^{SACA}_{i} +\textbf{MLP}(NL({S}^{SACA}_{i}))
\end{equation}
The outputs of FIA, S1, S2, S3,S4,S5 and S6 are then passed into the following CCM module.

\subsection{CCM: Channel agent cross attention module}

The post-interaction features obtained from the FIA stage cannot be directly upsampled and decoded.The CCM module can effectively connect the features extracted from the backbone network and the features processed by the cross-attention module using spatial agents, eliminating inconsistencies.As shown in Figure 6, first we downsample the encoder features to get tokens xi with the same size and number of channels as ei features.Then we conduct a pooling operation on tokens xi to get the Query Agent Tokens AQ.Agent Features are obtained by calculating the channel cross attention between agent feature AQ and feature ei.Similarly, we carry out the feature ei pooling operation to get the Key Agent Tokens AK.Feature xi, feature Query Agent Tokens and Agent Features are treated as Q,K and V respectively for the calculation of channel cross attention. Also we added Agent Bias and DWC operations.Eventually we reconstruct the resulting feature back to its original feature size.

\section{Experiments}
\subsection{Implementation Details}
We use Pytorch \cite{paszke2017automatic} to implement our model and one RTX 3090 GPUs are used for accelerating training. We resize images to 224×224 as input to Transformer and resize images to 1024×1024 as input to CNN.
We set the maximum learning rate to 0.003 for Swin backbone and 0.03 for others. The learning rate first increases then decays during the training process, what’s more Momentum and weight decay are set to 0.9 and 0.0005 respectively. Batchsize is set to 4 and maximum epoch is set to 50.For data
augmentation, we use random flip, crop and multi-scale input images\cite{qin2019basnet} \cite{tang2021disentangled}\cite{zhao2019egnet}.
\subsection{Datasets and Evaluation Metrics}
We follow recent methods to use the DUTS-TR (10553 images)\cite{wang2017learning}  to train our M3Net.We evaluate our M3Net on six widely used benchmark datasets, including DUT-OMORN \cite{yang2013saliency}  (5168 images),DUTS-TE \cite{wang2017learning} (5019 images), ECSSD \cite{yan2013hierarchical} (1000 images),HKU-IS \cite{li2015visual} (4447 images), PASCAL-S (850 images),
SOD \cite{movahedi2010design} (300 images).We use five metrics to evaluate our model and the existing state-of-the-art algorithms,namely MAE,E-measure\cite{fan2018enhanced} ,S-measure[fan2017structure],mF.

\subsection{Comparison with State-of-the-Art}
We compare our FIPGNet with 30 state-of-the-art methods,including U2Net, DASNet, F3Net, GateNet\cite{li2015visual},MINet \cite{pang2020multi}, LDF, PFS, CTD, VST, EDN, PGNet \cite{xie2022pyramid}, MENet.

\begin{figure}
\centering
\includegraphics[width=1\linewidth]{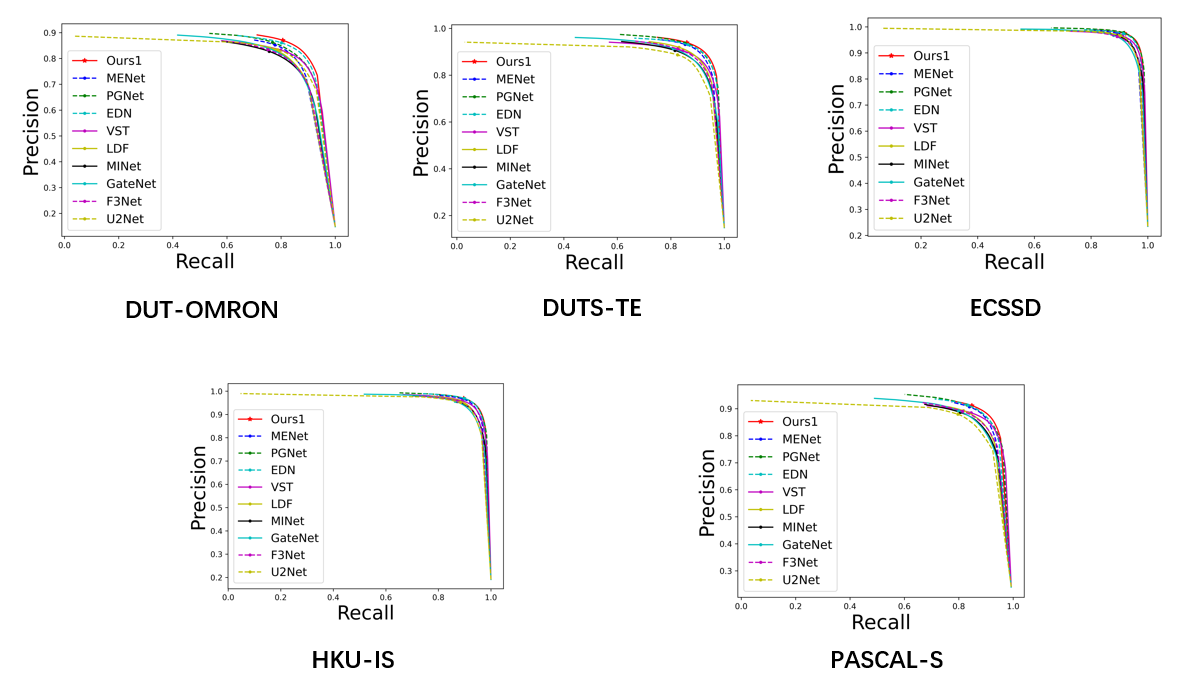}
\caption{Precision-Recall curves  of our FIPGNet and other ten SOTA methods on five benchmark datasets.
}
\label{fig:7}
\end{figure}

\begin{table*}[ht]
\centering
\renewcommand{\arraystretch}{1.0}
\setlength{\tabcolsep}{2pt}
\footnotesize
\caption{Performance Comparison. ResNet50 and T2T-ViT-14 are simplified to R50 and TV, respectively. ResNet50 with SwinB is simplified to RS.}
\begin{tabular}{lcccccccccccccc}
\toprule
Datasets/Methods & $U^2$Net & DASNet & $F^3$ Net & GateNet & MINet & LDF & PFS & CTD & VST & EDN & PGNet & MENet & Ours \\
\midrule
{Backbones}&RSU&R50&R50&R50&R50&R50&R50&R50&TV&TV&RS&R50 &RS\\\hline
MAE (DUTS-TE)     & .045 & .034 & .035 & .040 & .037 & .037 & .036 & .034 & .037 & .030 & .027 & .028 & .024 \\
MaxF (DUTS-TE)    & .873 & .896 & .891 & .887 & .883 & .897 & .898 & .897 & .890 & .914 & .917 & .912 & .920 \\
mF (DUTS-TE)      & .848 & .853 & .867 & .855 & .859 & .878 & .863 & .853 & .858 & .897 & .895 & .893 & .899 \\
mE$_m$ (DUTS-TE)  & .884 & .903 & .913 & .900 & .913 & .923 & .902 & .902 & .915 & .933 & .936 & .935 & .940 \\
S$_m$ (DUTS-TE)   & .874 & .894 & .888 & .885 & .884 & .892 & .894 & .894 & .896 & .918 & .917 & .917 & .918 \\
MAE (DUT-OMRON)   & .054 & .050 & .052 & .055 & .056 & .052 & .045 & .052 & .058 & .045 & .043 & .045 & .042 \\
MaxF (DUT-OMRON)  & .823 & .827 & .813 & .818 & .810 & .823 & .806 & .812 & .796 & .845 & .847 & .837 & .850 \\
mF (DUT-OMRON)    & .802 & .783 & .794 & .791 & .789 & .803 & .783 & .775 & .764 & .826 & .828 & .819 & .835 \\
mE$_m$ (DUT-OMRON)& .871 & .869 & .876 & .868 & .873 & .881 & .875 & .875 & .882 & .899 & .902 & .899 & .907 \\
S$_m$ (DUT-OMRON) & .846 & .845 & .838 & .838 & .833 & .839 & .842 & .844 & .840 & .871 & .873 & .866 & .870 \\
MAE (ECSSD)       & .033 & .032 & .033 & .041 & .034 & .030 & .031 & .033 & .033 & .029 & .026 & .027 & .022 \\
MaxF (ECSSD)      & .951 & .950 & .945 & .945 & .947 & .953 & .952 & .950 & .947 & .959 & .960 & .954 & .962 \\
mF (ECSSD)        & .932 & .932 & .934 & .925 & .931 & .938 & .930 & .925 & .927 & .948 & .948 & .936 & .946 \\
mE$_m$ (ECSSD)    & .925 & .940 & .946 & .943 & .953 & .951 & .928 & .926 & .925 & .961 & .952 & .938 & .968 \\
S$_m$ (ECSSD)     & .927 & .927 & .924 & .920 & .925 & .924 & .931 & .925 & .929 & .937 & .940 & .931 & .940 \\
\bottomrule
\end{tabular}
\end{table*}

\begin{figure}
\centering
\includegraphics[width=1\linewidth]{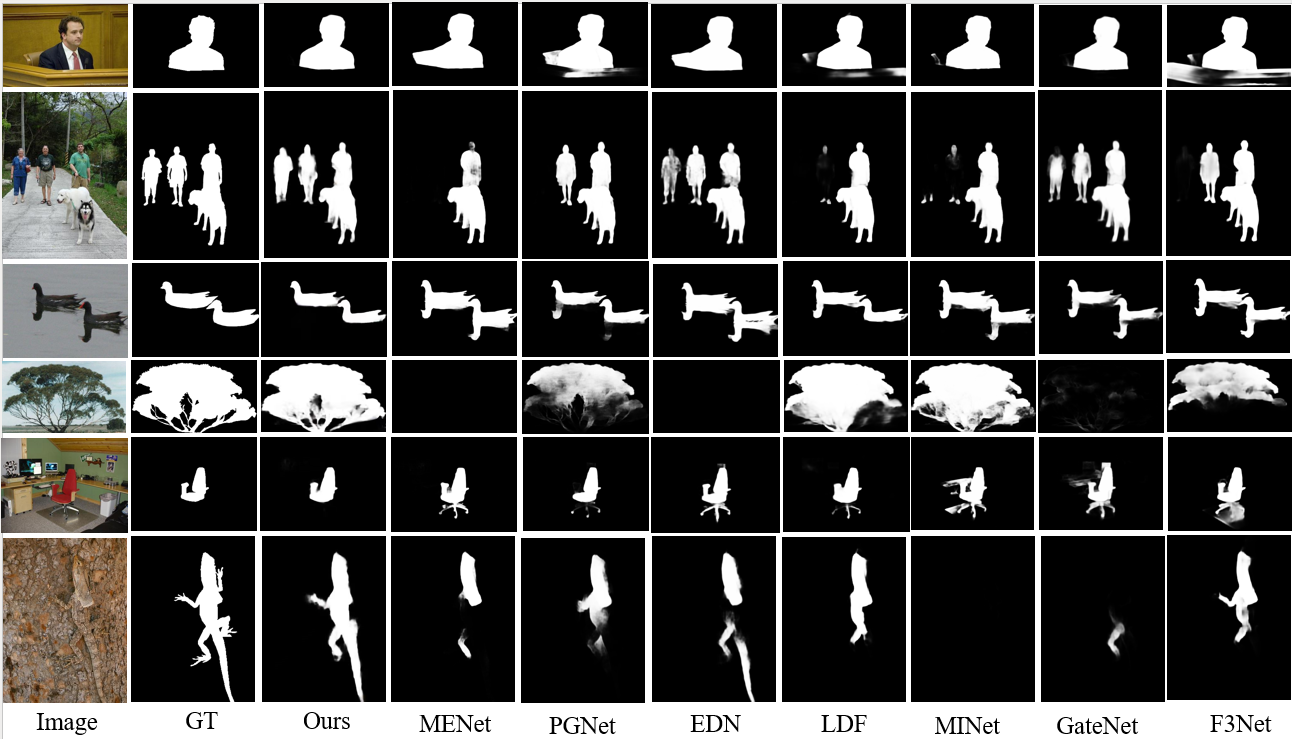}
\caption{Visual comparisons between our FIPGNet and other 7 state-of-the-art methods on various scenes}
\label{fig:8}
\end{figure}

\begin{table*}[ht]
\centering
\renewcommand{\arraystretch}{1.0}
\setlength{\tabcolsep}{2pt}
\caption{Performance Comparison. ResNet50 and T2T-ViT-14 are simplified to R50 and TV, respectively. ResNet50 with SwinB is simplified to RS.}
\footnotesize
\begin{tabular}{l|cccccccccccccc}
\toprule
Metrics/Methods & $U^2$Net & DASNet & $F^3$ Net & GateNet & MINet & LDF & PFS & CTD & VST & EDN & PGNet & MENet & Ours \\
\midrule
{Backbones}&RSU&R50&R50&R50&R50&R50&R50&R50&TV&TV&RS&R50 &RS\\\hline
MAE (HKU-IS)     & .031 & .027 & .028 & .034 & .029 & .027 & .027 & .027 & .030 & .024 & .024 & .023 & .021 \\
MaxF (HKU-IS)    & .935 & .942 & .936 & .933 & .935 & .939 & .937 & .945 & .942 & .951 & .948 & .948 & .951 \\
mF (HKU-IS)      & .913 & .917 & .918 & .909 & .916 & .922 & .923 & .920 & .921 & .938 & .933 & .932 & .940 \\
mE$_m$ (HKU-IS)  & .948 & .950 & .958 & .953 & .960 & .960 & .954 & .955 & .955 & .969 & .965 & .952 & .972 \\
S$_m$ (HKU-IS)   & .916 & .922 & .917 & .915 & .919 & .920 & .920 & .928 & .928 & .936 & .929 & .923 & .932 \\
MAE (PASCAL-S)   & .074 & .064 & .062 & .068 & .064 & .066 & .060 & .062 & .061 & .057 & .053 & .047 & .047 \\
MaxF (PASCAL-S)  & .859 & .885 & .871 & .869 & .866 & .864 & .874 & .875 & .857 & .893 & .892 & .896 & .850 \\
mF (PASCAL-S)    & .838 & .849 & .853 & .846 & .849 & .846 & .857 & .856 & .847 & .877 & .878 & .877 & .877 \\
mE$_m$ (PASCAL-S)& .850 & .872 & .894 & .884 & .898 & .884 & .861 & .875 & .861 & .911 & .911 & .920 & .855 \\
S$_m$ (PASCAL-S) & .844 & .860 & .861 & .858 & .856 & .857 & .863 & .861 & .861 & .877 & .878 & .881 & .850 \\
MAE (SOD)        & .106 & .098 & .092 & .098 & .092 & .093 & .086 & .093 & .086 & .083 & .094 & .085 & .085 \\
MaxF (SOD)       & .858 & .878 & .879 & .878 & .879 & .881 & .877 & .881 & .877 & .880 & .883 & .888 & .850 \\
mF (SOD)         & .843 & .875 & .871 & .875 & .871 & .865 & .862 & .865 & .862 & .869 & .867 & .881 & .877 \\
mE$_m$ (SOD)     & .799 & .870 & .870 & .870 & .870 & .866 & .856 & .866 & .856 & .841 & .822 & .855 & .855 \\
S$_m$ (SOD)      & .789 & .801 & .805 & .801 & .805 & .800 & .820 & .800 & .820 & .810 & .801 & .830 & .850 \\
\bottomrule
\end{tabular}
\end{table*}

\begin{table*}[ht] 
\centering 
\renewcommand{\arraystretch}{1.0}
\setlength{\tabcolsep}{2pt}
\caption{Performance comparison of different component settings across various datasets.} \label{table:performance} 
\footnotesize
\begin{tabular}{l|ccc|ccc|ccc|ccc|ccc} 
\hline 
Component Setting & \multicolumn{3}{c|}{DUSTS-TE} & \multicolumn{3}{c|}{DUT-OMRON} & \multicolumn{3}{c|}{ECSSD} & \multicolumn{3}{c|}{HKU-IS} & \multicolumn{3}{c}{PASCAL-S} \\
& MAE↓ & mE↑ & $S_m$↑ & MAE↓ & mE↑ & $S_m$↑ & MAE↓ & mE↑ & $S_m$↑ & MAE↓ & mE↑ & $S_m$↑ & MAE↓ & mE↑ & $S_m$↑ \\ \hline
Baseline & .027 & .922 & .909 & .045 & .887 & .855 & .028 & .952 & .936 & .024 & .965 & .929 & .053 & .908 & .877 \\ 
Baseline+FIA & .025 & .935 & .914 & .043 & .902 & .868 & .025 & .960 & .939 & .022 & .970 & .931 & .049 & .917 & .880 \\ 
Baseline+CCM & .025 & .936 & .915 & .045 & .904 & .869 & .023 & .962 & .939 & .023 & .968 & .930 & .049 & .918 & .881 \\ 
Baseline+MSCA+FAM & .024 & .940 & .918 & .042 & .907 & .870 & .022 & .968 & .940 & .021 & .972 & .932 & .047 & .920 & .881 \\ \hline 
\end{tabular} 

\end{table*}

\begin{itemize}
\item Quantitative Comparison:Table I shows the quantitative comparison results on six widely used benchmark datasets. We compare our method with 12 state-of-the-art methods in terms of MAE, E-measure,S-measure,mF.Figure 7 shows a comparison of the Precision-Recall curve between our method and nine other methods.As can be seen from the figure, our method achieves the best results on the DUT-OMRON,DUTS-E,PASCAL-S datasets, as well as competitive results on the ECSSD and HKU-IS datasets.
.
\item Visual Comparison:Figure 8 provides visual comparisons between our method and other methods.We compared our method with seven other methods on single-person images, multi-person images, reflection images in water, outdoor landscape images, indoor complex images, and low-contrast images.
As can be seen, our FIPGNet reconstructs more accurate saliency maps.The above results show the versatility and robustness of our FIPGNet.
\end{itemize}

\subsection{Ablation Studies}
To demonstrate the effectiveness of different modules in
our FIPGNet, we conduct the quantitative results of our method. The experimental results on three datasets including DUTS-TE, ECSSD, HKU-IS ,DUT-OMRON ,PASCAL-S are given in Table 3.We start with the pyramidal graft network structure and gradually add our proposed modules, including FIA and CCM.

\section{Conclusion}
In this paper, we propose a novel Transformer based network dubbed FIPGNet for SOD.Considering the spatial correlation between features, we first propose a feature interaction strategy (FIA) based on attention mechanism to realize multi-level feature interaction.Among them, we innovatively introduce the spatial agent Cross Attention (SACA) block.Finally, we propose a channel agent cross attention module (CCM) for efficiently connecting features extracted from the backbone network with features processed using the spatial agent cross attention module.The experimental results show that our model, FIPGNet, achieved competitive results on six datasets, demonstrating the great potential of our model in the SOD mission.

\bibliographystyle{unsrt}
\bibliography{sample}

\end{document}